# AnyOKP: One-Shot and Instance-Aware Object Keypoint Extraction with Pretrained ViT

Fangbo Qin, Taogang Hou, Shan Lin, Kaiyuan Wang, Michael C. Yip, Shan Yu

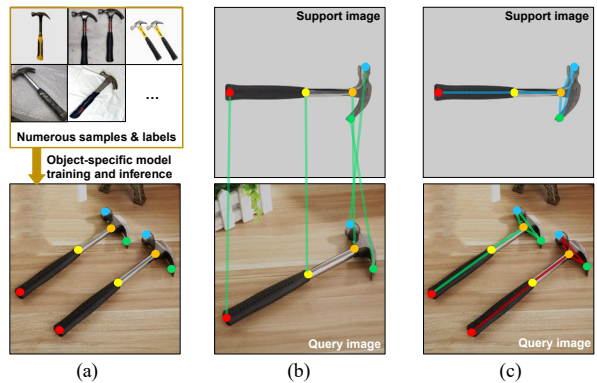

Fig. 1. Three object keypoint extraction paradigms. (a) Object-specifc extraction. (b) One-shot and instance-unaware extraction by one-to-one correspondence. (c) One-shot and instance-aware extraction (Proposed).

*Abstract*—Towards flexible object-centric visual perception, we propose a one-shot instance-aware object keypoint (OKP) extraction approach, AnyOKP, which leverages the powerful representation ability of pretrained vision transformer (ViT), and can obtain keypoints on multiple object instances of arbitrary category after learning from a support image. An off-the-shelf petrained ViT is directly deployed for generalizable and transferable feature extraction, which is followed by training-free feature enhancement. The best-prototype pairs (BPPs) are searched for in support and query images based on appearance similarity, to yield instance-unaware candidate keypoints. Then, the entire graph with all candidate keypoints as vertices are divided to sub-graphs according to the feature distributions on the graph edges. Finally, each sub-graph represents an object instance. AnyOKP is evaluated on real object images collected with the cameras of a robot arm, a mobile robot, and a surgical robot, which not only demonstrates the cross-category flexibility and instance awareness, but also show remarkable robustness to domain shift and viewpoint change.

## I. INTRODUCTION

Object-centric visual perception is widely used in robotic grasping, assembly, searching, surgery, etc. [1-4].

Keypoints, a set of identifiable points of interest on an object's surface, are preferred in object perception because they not only have semantic information of object details, but also indicate the object's spatial and structural information. Besides, sparse keypoints benefit the computation efficiency.

In recent years, object keypoint extraction methods based on deep neural networks (DNNs) have been applied to a variety of robot and automation tasks. For robot manipulation, keypoints were detected to represent the affordance geometry [5]. For robotc door opening, three semantic keypoints were used as the compact representation of door handle [6]. Point cloud keypoints of tools were used for robotic manipulation [7]. Pose estimation of objects could be based on keypoints [8]. In robotic surgical vision, keypoint extraction could be used to localize surgical tools [9, 10]. The above methods are all *object-specific*, in which the models are trained and tested on the same object categories, as shown in Fig. 1(a). Given a novel object category, the model has to be re-trained or adapted with non-negligible costs on time and labor.

For flexible deployment, *object-agnostic* visual perception can work on previously unseen object category with minor cost on time and labor. A typical and effective paradigm is based on the appearance consistency of object over different scenes and imaging conditions. To perceive an object in an unknown *query image*, a model can refer to a known *support image* containing the same object category and perform the feature-consistency based correspondence between the support and query images. This paradigm is called *one-shot learning* if the model explicitly learns the storable *prototype*s (also called exemplars) from support image [11-14]. The point correspondence methods also can be regarded as one-shot learning if we deem one of the paired images as support image and the other as query image [15-17].

We notice two limitations in one-shot keypoint extraction: 1) It is difficult to learn a representation that not only can discriminate various object details but also keep invariant under significant imaging condition changes. A main reason is the lack of large dataset specially for one-shot learning task. Compared to many category-oriented large datasets whose image samples are deemed individual to each other, the dataset for one-shot learning requires paired images that have correspondences in contents. Because paired images are difficult to collect and label, current self-supervised training methods usually use random homography transformation [18-20] and structure from motion [21,22] to generate point correspondences in paired images. However, the diversity of real object categories and background scenes are limited. 2) Current methods only consider the one-to-one correspondence, namely each keypoint in support image corresponds to

This work was supported by the National Key Research and Development Program of China (2021ZD0200402), the National Natural Science Foundation of China (62103413,62103035), and the Young Elite Scientists Sponsorship Program by CAST (2022QNRC001).

F. Qin and S. Yu are with the Institute of Automation, Chinese Academy of Sciences, Beijing 100190, China. T. Hou is with the School of Electronic and Information Engineering, Beijing Jiaotong University, Beijing, 100044, China. S. Lin, K. Wang, and M.C. Yip are with the Department of Electrical and Computer Engineering, University of California San Diego, La Jolla, CA 92093, USA. {qinfangbo2013@ia.ac.cn}

only one keypoint in query image, as shown in Fig. 1(b). Without relying on hard thresholding, max-out operation simply selects the *unilaterally* best matching pixel [20], and the best buddy pair (BBP) method select the *mutually* best matching pixel for better robustness [23,24], which usually work in structured scenes or in 3D reconstruction task. However, when multiple object instances occur in a query image, such as the two hammers in Fig. 1(c), neither max-out nor BBP extraction can obtain duplicated keypoints and attach them to different object instances.

Considering these two issues, we propose a one-shot and instance aware keypoint extraction approach. 1) Thanks to the cutting-edge ViT models pretrained on super large datasets, we exploit the off-the-shelf ViT features for transferable and generalizable keypoint description. 2) A best-prototype pair (BPP) extraction method is proposed to select candidate keypoints from query image, which permits one-to-many correspondence so that keypoints on multiple object instances can be collected. 3) To realize instance-awareness, we propose an edge-based keypoint grouping method to divide an graph with all the candidate keypoints as vertices to sub-graphs, by pruning the invalid graph edges with less consistent feature distributions. 4) The proposed AnyOKP approach is comprehensively evaluated on various 3D objects in challenging scenes with three typical robots.

## II. PRELIMINARIES

### A. Pretrained ViT

After convolutional neural networks (CNNs), typified by ResNet [25], dominated the area of computer vision for years, recently the Transformer based models, typified by ViT [26], emerged as a powerful alternative to CNNs. As reported in [27], ViTs possess the stronger robustness against texture change, partial occlusion, domain shift, and image corruptions than CNNs. After pretraining on super large dataset, ViTs outperform CNNs and provide off-the-shelf transferable visual features for downstream tasks like image recognition and segmentation [26, 28]. The merits of ViTs intrigued researchers in robotics and automation areas [11,29,30].

CNNs excel at learn texture patterns and local shapes, due to the hierarchical convolution and pooling operations. ViTs are based on the self-attention mechanism, which regard image patches as tokens, then capture their semantics and global relations with $L$ Transformer layers. As shown in Fig. 2, an input image $\mathcal{I}$ is divided to $P$ patches with the stride $s$ and patch size $p \times p$. Each patch is flattened to a vector with $p^2 \times 3$ dimensions, then transformed to a $D$-dimensional patch embedding $x_i$ ($i=1,2,…,P$) with linear projection. The patch embeddings and an extra $D$-dimensional vector $c$ for image description are concatenated, then added with position embeddings $r_i$. Thus, we have the $P+1$ tokens,

$$t = \{c+r_0, x_1+r_1, x_2+r_2, …, x_P+r_P\}. \quad (1)$$

The tokens are fed to $L$ cascaded Transformer layers. Each Transformer layer updates the tokens by

$$t'_l = \text{MHSA}(\text{LN}(t_{l-1})) + t_{l-1}, \quad t_l = \text{MLP}(\text{LN}(t'_l)) + t'_l, \quad (2)$$

where $l$ is the layer index. MHSA, MLP, and LN stand for multi-head self-attention [31], multilayer perceptron, and layer normalization, respectively. The feature dimension $D$ is constant over all the layers. The final tokens $t_L$ of the $L^{\text{th}}$ layer are the patch embeddings $\{z_i\}$ ($i=1,2,…,P$), which can be reorganized as a feature map $\mathcal{Z}$ with $P$ pixels and $D$ channels.

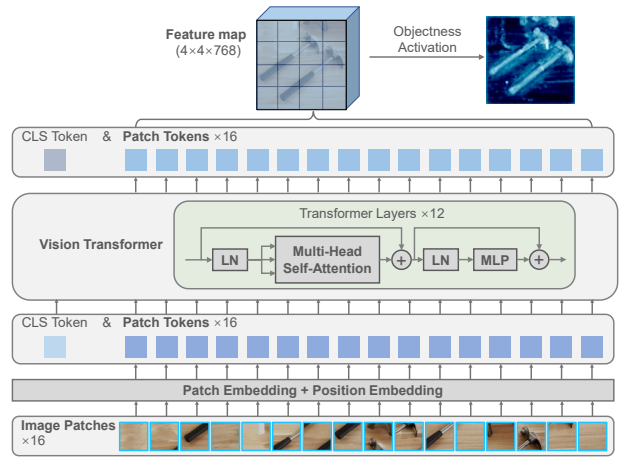

Fig. 2. Vision Transformer as feature extractor. As a brief example, the input image is divided to $P$=16 (4×4) patches, and the feature dimension $D$ is 768.

The ViT's performance outperforms CNN only when the pretraining dataset is large enough, because ViT has to learn the translation equivariance and neighborhood structure from scratch, while CNN owns these two properties inherently. Aside from the supervised training on labeled dataset [26,32], the self-supervised training on unlabeled dataset not only achieves competitive performance but also demonstrate emerging properties: *understanding of object parts and scene geometry regardless of the image domains* [33,34].

### B. Keypoint Similarity Measure

The correspondence and distinction of object keypoints are based on feature descriptors given by CNN [15-18, 20-22] or ViT[23]. With ViT as the feature extractor, a keypoint lying in the $i^{\text{th}}$ image patch is described by the $i^{\text{th}}$ patch embedding $z_i$. Cosine similarity $\psi$ is used to measure the similarity between keypoints $i$ and $j$, namely,

$$\psi(z_i, z_j) = (z_i \cdot z_j) / (\|z_i\|_2 \cdot \|z_j\|_2). \quad (3)$$

Since keypoint descriptor is not ideally invariant across scenes and imaging conditions, the similarity between corresponding keypoints might be pulled down by significant changes of background, pose, lighting, *etc*. Therefore, the one-shot keypoint extraction cannot be simply realized by setting a hard threshold and selecting similar points.

## III. ONE-SHOT INSTANCE-AWARE KEYPOINT EXTRACTION

### A. Training-Free Feature Enhancement

Although the pretrained ViT features are off-the-shelf and transferable, it is essential to enhance the features in a training-free manner for a decent performance.

*1) Objectness Attention*: The mean absolute activation of each pixel is calculated by,

$$\mathcal{O}_i = \frac{1}{D} \sum_j |z_{i,j}|, \quad (4)$$

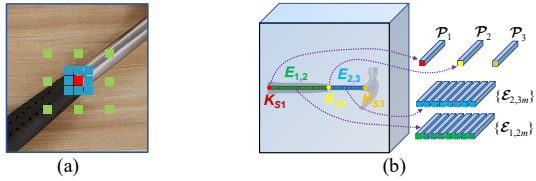

Fig. 3. Illustrations of (a) neighborhood binning and (b) learning of keypoint prototypes and edge prototypes from support image.

where $z_{i,j}$ is the feature value of patch $i$ and channel $j$. The activation map $\mathcal{O}$ is transformed to the range [-1,1] by min-max normalization. We found the raw DINO-ViT features [33] have strong activation on objects and low activation on meaningless background, as shown in Fig. 2. Therefore, we call $\mathcal{O}$ objectness activation map. In comparison, SAM-ViT [32] and DINOv2-ViT [34] have weaker object awareness when using (4). The former has strong activation only on contours and textures, the latter has unclear activation. CNNs including LOFTR FPN [22] and DINO ResNet50 [33] hardly present any objectness awareness when using (4).

The objectness attention mechanism is realized by scaling the feature map in a pixel-wise manner,

$$\mathcal{Z}_{Ai} = \sigma(\alpha \mathcal{O}_i) \times \mathcal{Z}_i. \quad (5)$$

Where $\sigma$ is the sigmoid function and the scaling factor $\alpha$ is set as 5. Thus, the features of the pixels with low objectness activation are further suppressed.

2) *Neighborhood Binning*: Following [23], we use binning operation to incorporate the features of each pixel and those of its neighboring pixels. Firstly, a pooled feature map $\mathcal{Z}_{POOL}$ is obtained by average pooling on $\mathcal{Z}_A$ with stride 1 and kernel size 3. Secondly, for each pixel we collected its eight adjacent pixels' features from $\mathcal{Z}_A$ and its eight spaced adjacent pixels' features from $\mathcal{Z}_{POOL}$, where the spacing is 3 pixels, as shown in Fig. 3(a). The collected features are concatenated with the center pixel's features. Thus, $\mathcal{Z}_A$ with $D$ channels is enhanced to $\mathcal{Z}_B$ with $D_B=17 \times D$ channels. Although more channels lead to heavier computation load, this step is essential because ViT does not excel at encoding 2D local patterns and the neighborhood binning can enhance the local structural description by organizing the semantics nearby a keypoint.

B. *One-Shot Prototypes Learning*

In the one-shot keypoint extraction paradigm, we use an image $\mathcal{I}_S$ containing a single target object as the support sample, to provide priori information of object and keypoints. $N_{KP}$ keypoints are designated on the target object, which are denoted as $\{K_{Sk}\}$ ($k=1,2,…,N_{KP}$). $k$ is regarded as the *keypoint identity*. A pretrained ViT and the feature enhancement modules are used to convert $\mathcal{I}_S$ to a feature map $\mathcal{Z}_{SB}$ that has $P$ pixels and $D_B$ channels.

For each keypoint $K_{Sk}$, its coordinates $(u_{Sk}, v_{Sk})$ on $\mathcal{I}_S$ is mapped to the index $i_{Sk}$ on $\mathcal{Z}_{SB}$. The $D_B$-dimensional feature vector $\mathcal{P}_k$ is withdrawn from $\mathcal{Z}_{SB}$ with the index $i_{Sk}$, and is called a *keypoint prototype*. The keypoint prototypes $\{\mathcal{P}_k\}$ ($k=1,2,…,N_{KP}$) are used in the following BPP extraction step.

To prepare for keypoint grouping, the feature distributions between keypoints are also learned from $\mathcal{I}_S$. As shown in Fig. 3(b), regarding keypoints as vertices and an instance as a graph, the graph edges represent the connectivity within an instance. To describe the edge $E_{kl}$ linking the keypoints with the identities $k$ and $l$, the edge is evenly divided to $N_{SEG}=8$ sub-segments. The features on each sub-segment are averaged as a descriptor $\mathcal{E}_{klm}$. Finally, the set of descriptors $\{\mathcal{E}_{klm}\}$ ($m=1,2,…,N_{SEG}$) is used as the *edge prototype* for $E_{kl}$. Note that for objects with non-convex or skeleton-like shapes, the edges lying out of object foreground are not used.

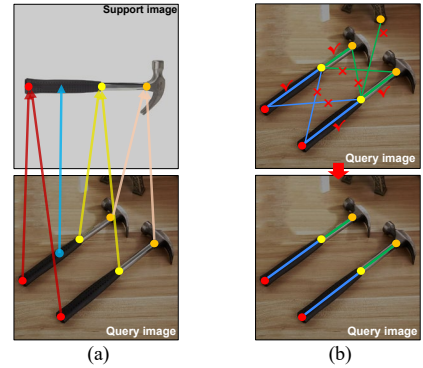

Fig. 4. BPP extraction and keypoint grouping. (a) Seven BPPs are depicted as examples. The BPP shown in blue is rejected because its best prototype is not a required keypoint. (b) Graph vertices and edges in query image. The color of keypoint indicates identity. The width of edge indicates the edge similarity given by (8). The invalid edges in the initial graph are pruned.

C. *BPP Extraction*

Given a query image $\mathcal{I}_Q$, the object instances and keypoints are all unknown. Similar to $\mathcal{Z}_{SB}$, the shared pretrained ViT and feature enhancement modules are used to convert $\mathcal{I}_Q$ to a feature map $\mathcal{Z}_{QB}$ that has $P$ elements and $D_B$ channels. Then, the similarity matrix $\mathcal{S}$ is calculated by,

$$\mathcal{S}_{ij} = \psi(\mathcal{Z}_{SBi}, \mathcal{Z}_{QBj}), \quad (6)$$

where $\mathcal{S}_{ij}$ indicates the similarity between the $i^{th}$ pixel in $\mathcal{Z}_{SB}$ with the $j^{th}$ pixel in $\mathcal{Z}_{QB}$. For each query pixel $j$ in $\mathcal{Z}_{QB}$, its *best prototype* is indexed by,

$$i_{BPj} = \arg\max_i \mathcal{S}_{ij}. \quad (7)$$

Thus, the query pixel $j$ in $\mathcal{Z}_{QB}$ and the support pixel $i_{BPj}$ in $\mathcal{Z}_{SB}$ form a BPP. As shown in Fig. 4(a), for each keypoint $K_{Sk}$ in $\mathcal{I}_S$, all the pixels in $\mathcal{I}_Q$ that form BPPs with $K_{Sk}$ are gathered as candidate keypoints $\{K_{Ckn}\}$ ($n=1,2,…,N_{Ck}$) with the identity $k$. For better robustness, the query pixels forming BPPs with the four adjacent pixels around $K_{Sk}$ are also collected. Finally, non-maximum suppression (NMS) with a weak threshold $\tau_{KP}=0.0$ is applied to the candidate keypoints, so that the redundant and unconfident candidates are removed.

D. *Edge-Based Keypoint Grouping*

As shown in Fig. 4(b), the candidate keypoints $\{K_{Ckn}\}$ ($k=1,2,…,N_{KP}$; $n=1,2,…,N_{Ck}$) in $\mathcal{I}_Q$ form the vertices of an initial graph. Keypoint grouping is to divide the initial graph to subgraphs and each subgraph represents an object instance. For a candidate edge linking two keypoints with the identities $k$ and $l$, its descriptor $\{\mathcal{E}_{Cklm}\}$ ($m=1,2,…,N_{SEG}$) is obtained in the same manner of obtaining $\{\mathcal{E}_{klm}\}$. The edge similarity between $\{\mathcal{E}_{Cklm}\}$ and $\{\mathcal{E}_{klm}\}$ is measured by,

$$\phi(\{\mathcal{E}_{Cklm}\}, \{\mathcal{E}_{klm}\}) = \frac{1}{N_{SEG}} \sum_m \psi(\mathcal{E}_{Cklm}, \mathcal{E}_{klm}). \quad (8)$$

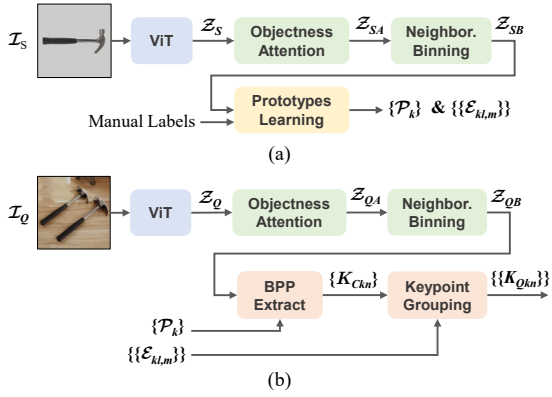

Fig. 5. AnyOKP pipeline. (a) Offline one-shot prototypes learning with support image and manual annotation. (b) Online instance-aware keypoint extraction with query image and learned prototypes.

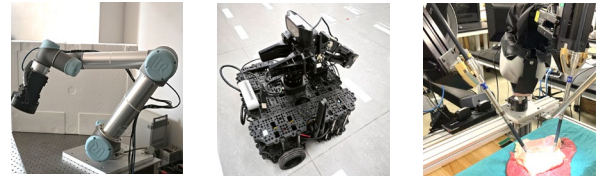

Fig. 6. Robot platforms for query image collection.

The default $N_{SEG}$ is 8. Thus, (8) evaluates the average similarity between the $N_{SEG}$ segments of a candidate edge and those of its prototype edge.

As shown in Fig. 4(b), in an initial graph, a valid edge connects two keypoints on the same object instance, which is expected to have a high edge similarity score. An invalid edge connects two keypoint on two different object instances or connects a keypoint and a false positive one, which is expected to have a low edge similarity score. Except edges linking two false positives, an invalid edge shares a true positive vertex with a valid edge. To prune invalid edges, firstly a weak threshold $\tau_E=0.3$ is used to reject edges with excessive low similarities. Afterwards, for each two edges sharing a vertex, if their unshared two vertices also have the same keypoint identity, then the one with the lower edge similarity is pruned. Finally, after pruning the invalid edges misconnecting two different instances, the initial graph is divided to individual subgraphs, each of which represent an instance. The keypoint groups $\{\{K_{Qkn}\}\}$ are gained from the subgraphs. $n$ is the instance index and $k$ is the keypoint identity.

### E. AnyOKP Pipeline

The pipeline of the proposed approach is shown in Fig. 5. The raw image is resized to the required input size of ViT model. The learning of keypoint prototypes and edge prototypes is conducted beforehand which requires the human's annotations. The learned prototypes are stored and used in the following online query image perception tasks. The post-process steps include: 1) scale the coordinates to align with the raw image's size, 2) remove the unreliable object instances whose valid keypoint numbers are lower than the minimum number $\tau_{KP}$. When $N_{KP} \leq 4$, $\tau_{KP}$ is set as max(2, $N_{KP}-1$). When $N_{KP}>4$, $\tau_{KP}$ is set as 4, because four keypoints that are not colinear already can reliably represent a rigid object's pose. For example, given the query image in Fig. 1(c), if one of the five keypoints is missed or occluded, the object instance can still be effectively localized.

## IV. EXPERIMENTS

### A. Evaluation Dataset and Metrics

We built a challenging dataset featured by two difficulties: 1) *Domain Shift between support and query images*: For each target object, its support image was captured by a cellphone camera in a clean scene, while its query images were captured by a robot camera in a different complex scene. 2) *Viewpoint Changes in query image sequence:* The perspective transformation of 2D object shape is noticeable in query images due to the viewpoint changes around the 3D object.

The robots used for query image collection are shown in Fig.6. First, a Basler acA2440 industrial camera on a UR5 robot's end was used to capture four image sequences in table scenes, with *watering can*, *paper box*, *screw driver*, and *electric drill* as target objects, respectively. Second, a Logitech C920 camera on the end of a Turtlebot3 mobile robot's lightweight arm was used to capture two image sequences in room scenes, with *fire extinguisher* and *trolley* as target objects, respectively. Third, endoscopic camera on a dVRK surgical robot's body was used to capture two image sequences in pseudo surgical scenes, with *needle driver* and *irrigator* as target objects, respectively. In total, eight query image sequences were collected and their ground truths (GTs) were labeled. Each sequence had 50-80 query images. All the images were padded to square and resized to 512×512.

For the extraction result of each query image, we calculated the *keypoint recall* ($R_{KP}$), *keypoint precision* ($P_{KP}$), *instance recall* ($R_{INS}$), and *instance precision* ($P_{INS}$). An extracted keypoint was deemed a true positive (TP) keypoint if its distance to GT was lower than 5% of the image width. An extracted object instance was considered a TP instance when it owned no fewer than $\tau_{KP}$ TP keypoints and no more than $N_{KP}$ TP keypoints. $\tau_{KP}$ is introduced in Section III.*E*. Then for each object category, the average scores $\bar{R}_{KP}$, $\bar{P}_{KP}$, $\bar{R}_{INS}$, and $\bar{P}_{INS}$ were calculated over all images in a query image sequence. Finally, for the whole evaluation dataset, the mean scores of $\bar{R}_{KP}$, $\bar{P}_{KP}$, $\bar{R}_{INS}$, and $\bar{P}_{INS}$ over the eight object categories were calculated to reflect the overall performances.

### B. Implementation Details

The ViTs and CNNs used in the experiments were all off-the-shelf and downloaded from Github repositories. Therefore, the patch size $p$ of ViT and the inner structure of CNN could not be changed. Firstly, we selected a uniform resolution of 64×64 for the feature map $\mathcal{Z}$, as a balance between spatial resolution and computation load. For ViTs [32-34], the strides were uniformly set as $p/2$ and the input sizes were accordingly set to guarantee the 64×64 resolution of $\mathcal{Z}$. For example, the input sizes of ViTb-$p$8 and ViTb-$p$16 were set as 260×260 and 520×520, respectively. For CNNs, the input size was uniformly 512×512. The coarse-level output of FPN in LoFTR [22] was a 64×64 feature map. The final output of ResNet50 was a 64×64 feature map when we changed the strides of the last two blocks from 2 to 1. Thus, all the ViTs and CNNs could be plugged into the same AnyOKP pipeline and be fairly compared. The computation

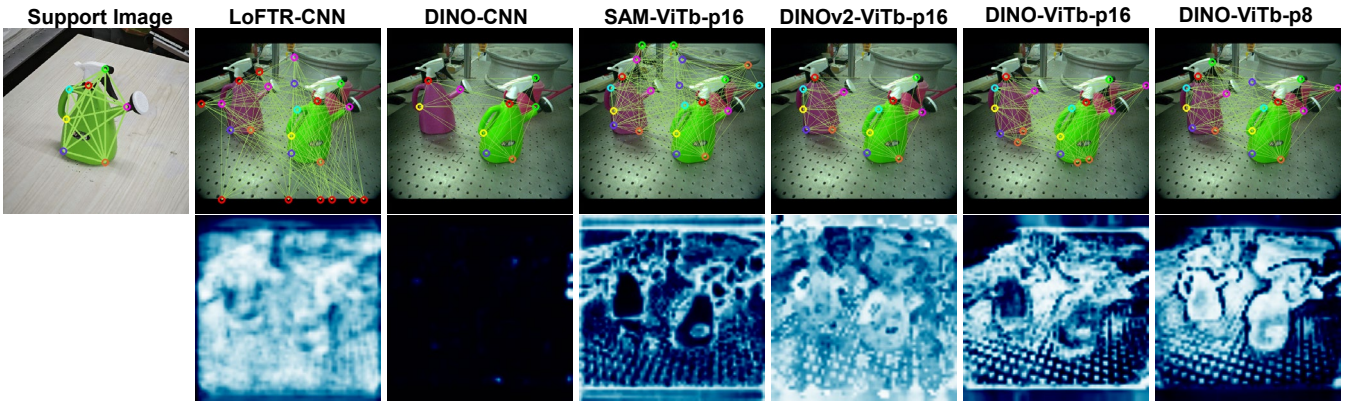

Fig. 7. Candidate keypoints and activation maps. The 2nd to 7th columns visualize the results with different feature extractors. In each colomn, the upper subfigure shows the extracted candidate keypoints (circles) and the initial graph (green lines) in the query image, while the lower subfigure visualizes the activation map given by (4), min-max normalization, and sigmoid function. Note that only a weak threshold $\tau_{KP}=0.0$ and the NMS operation are used to filter the candidate keypoints and the remaining false positives can be further rejected in the following grouping step. The color of keypoint indicates the identity.

TABLE I
BPP BASED CANDIDATE KEYPOINT EXTRACTION PERFORMANCES

| Feature Extractor[1] | Feature Enhance | Mean $\bar{R}_{KP}$ | Mean $\bar{P}_{KP}$ |
|---|---|---|---|
| LoFTR-**CNN** [22] | × | 0.644 | 0.372 |
| | NB | 0.723 | 0.463 |
| DINO-**CNN** [33] | × | 0.222 | 0.230 |
| | NB | 0.450 | 0.519 |
| SAM-**ViTb**-p16 [32] | × | 0.838 | 0.479 |
| | NB | 0.906 | 0.668 |
| DINOv2-**ViTb**-p14 [34] | × | 0.746 | 0.634 |
| | NB | 0.924 | 0.849 |
| DINO-**ViTb**-p16 [33] | × | 0.636 | 0.459 |
| | NB | 0.881 | 0.883 |
| DINO-**ViTb**-p8 [33] | × | 0.717 | 0.529 |
| | NB | 0.948 | **0.887** |
| | NB+OA | **0.961** | 0.885 |

[1] p16, p14, and p8 stand for patch sizes 16×16, 14×14, and 8×8, respectively.

TABLE II
ONE-SHOT AND INSTANCE-AWARE KEYPOINT EXTRACTION PERFORMANCE

| Object Category | $N_{KP}$ | $N_{INS}$ | Object Keypoint | | Object Instance | |
|---|---|---|---|---|---|---|
| | | | $\bar{R}_{KP}$ | $\bar{P}_{KP}$ | $\bar{R}_{INS}$ | $\bar{P}_{INS}$ |
| Watering can | 7 | 2 | 0.953 | 0.990 | 1.000 | 1.000 |
| Paper box | 4 | 4 | 0.980 | 0.997 | 0.991 | 1.000 |
| Screwdriver | 4 | 4 | 0.905 | 1.000 | 0.930 | 1.000 |
| Electric drill | 5 | 1 | 0.964 | 0.964 | 1.000 | 1.000 |
| Extinguisher | 2 | 2 | 0.982 | 0.982 | 0.965 | 0.965 |
| Trolley | 8 | 2 | 0.867 | 0.965 | 0.974 | 0.974 |
| Needle driver | 2 | 2 | 0.878 | 0.970 | 0.849 | 0.941 |
| Irrigator | 2 | 1 | 1.000 | 1.000 | 1.000 | 1.000 |
| *Mean* | / | / | *0.941* | *0.984* | *0.964* | *0.985* |

was based on a Intel Xeon Silver 4214R CPU and four Nvidia Geforce RTX3090 GPUs.

### C. Comparison Experiments

The BPP based candidate keypoint extraction is the basis of one-shot and instance-aware keypoint extraction. If many false positives appear or many keypoints are missed, the final result of AnyOKP will be irreparable. Therefore, we conducted a series of BPP based candidate keypoint extraction experiments without involving the keypoint grouping step. The different feature extractors realized by ViTs and CNNs [22, 32-34] were compared, to investigate which pretrained model provided the best feature representation in the keypoint extraction task. Besides, the influences of the feature enhancement modules neighborhood binning (NB) and objectness attention (OA) were also studied. The comparison results are reported in Table I and visualized in Fig. 7.

*1) Effectiveness of NB*: The neighborhood binning could significantly improve both the recall and precision scores of all the feature extractors.

*2) ViTs vs. CNNs*: The CNNs' performances were much more inferior than ViTs' when facing significant domain shift and viewpoint change. As shown in Fig. 7, LoFTR-CNN and DINO-CNN both could extract all or most TP keypoints on the right green water can, but could not recall enough TP keypoints on the left red water can. In comparison, all ViTs could recall 6-7 out of 7 keypoints on both two water cans, presenting the better generalization ability.

*3) Best ViT Version*: Among the four ViTs, DINO-ViTb-p8 had the best performance. DINO-ViTb-p16 had the much lower recall, although it differed with DINO-ViTb-p8 only on patch size. We tried to adjust the input size and stride of DINO-ViTb-p16 but still could not achieve a performance competitive to that of DINO-ViTb-p8. Therefore, a reason that DINO-ViTb-p8 outperformed the other three ViTs was the smaller patch size. A larger patch size like 14×14 or 16×16 involving much more pixels might lead to weaker generalization ability for keypoint description. Because the DINOv2 and SAM works did not release off-the-shelf ViT versions with smaller patch sizes, DINO-ViTb-p8 was the only available version with a small patch size in this work.

*4) Effectiveness of OA*: When using the proposed (4), min-max normalization, and sigmoid function to obtain activation maps, we found an interesting phenomenon that the DINO-ViTb-p8 pretrained on large dataset with no labels presented remarkable awareness on object foreground, as shown Fig. 7. In comparison, CNNs hardly showed any objectness awareness, and the other ViTs showed weaker objectness awareness, with our brief method. Therefore, the objectness attention mechanism in (5) was only suitable for DINO-ViTb-p8, which indeed improved the mean $\bar{R}_{KP}$ by 0.013 while the mean $\bar{P}_{KP}$ was slightly decreased by 0.002.

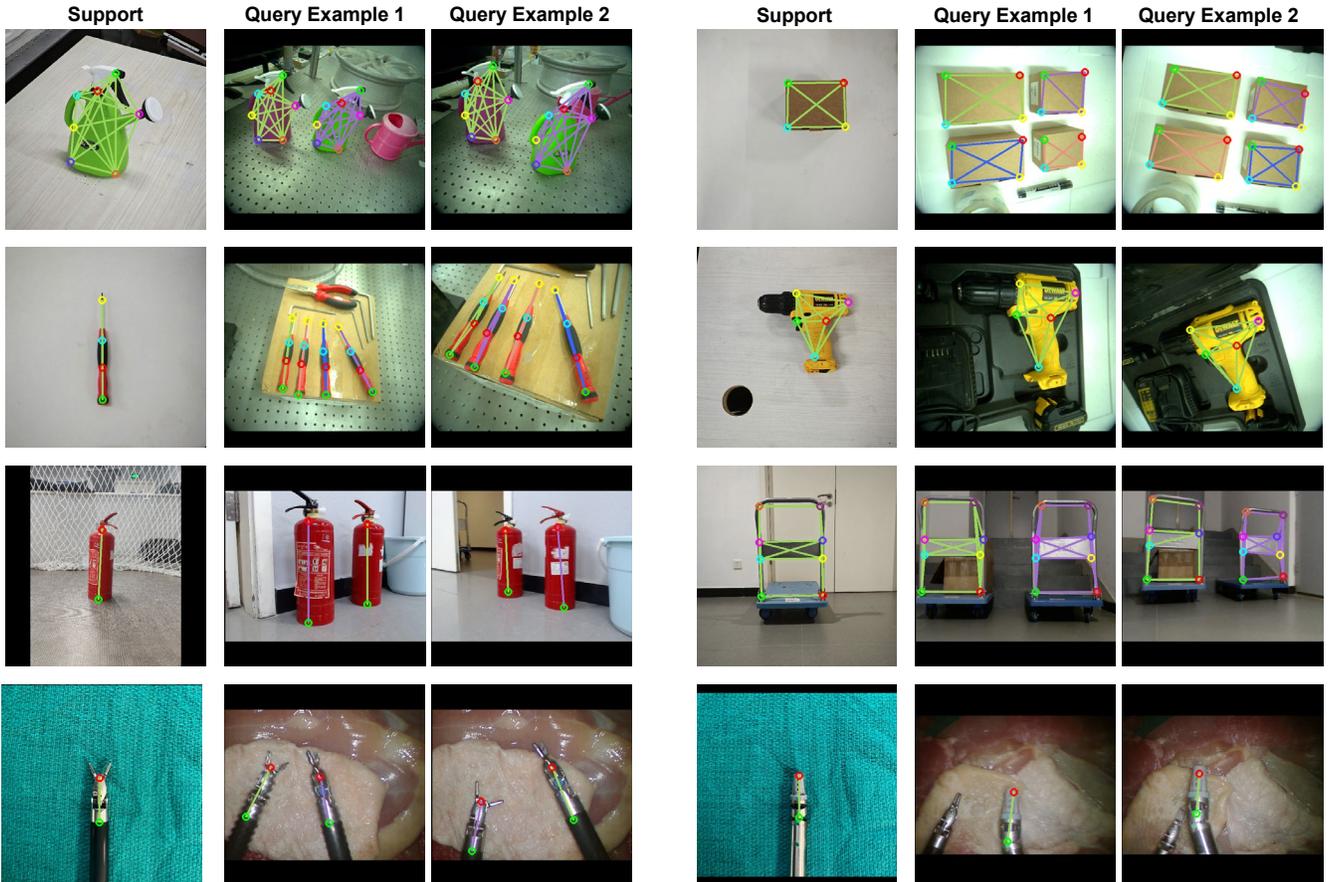

Fig. 8. One-shot and instance-aware keypoint extraction results with DINO-ViTb-p8 as feature extractor. Each circle marks a keypoint and its color indicate the keypoint identity. Each subgraph formed by line-segment edges represent an object instance. Different colors of line segments indicate different instances.

*D. One-Shot and Instance-Aware Keypoint Extraction*

In this experiment, we adopted the DINO-ViTb-p8 as the feature extractor, and run the whole AnyOKP pipeline on the evaluation dataset. Specially, when processing the query images collected with surgical robot, the threshold $\tau_E$ was adjusted from 0.3 to 0.0 empirically. Table II shows the scores $\bar{R}_{KP}$, $\bar{P}_{KP}$, $\bar{R}_{INS}$, and $\bar{P}_{INS}$ for each of the eight object categories as well as the mean scores over eight categories. First, comparing to Table I, the mean $\bar{P}_{KP}$ was significantly improved from 0.885 to 0.984, while the mean $\bar{R}_{KP}$ was decreased from 0.961 to 0.941, because the keypoint grouping step could reject isolated false positive keypoints and unreliable keypoints forming invalid subgraphs. Second, as shown by the object instance recalls and precisions, the proposed AnyOKP demonstrated the feasibility and reliability to identify one or multiple object instances based on their keypoints.

Some extraction results are visualized in Fig. 8, which intuitively demonstrate three advantages: 1) AnyOKP could be applied to various different object categories and the one-shot learning with support image was convenient. 2) AnyOKP not only could extract the keypoints with semantics, but also could identify different object instances. 3) AnyOKP showed remarkable robustness to domain shift and viewpoint change. Besides, for rigid-body object, even when a small part of keypoints are missed, the object can still be localized by the other extracted keypoints.

*E. Limitations*

To the best of our knowledge, AnyOKP is the first one-shot keypoint extraction method that can deal with multiple object instances. However, the current version has the following limitations: 1) None real-time: the forwarding of ViTb already costs ~148ms, and the neighborhood binning step costs over one second because the binning operator in the current version is not implemented with CUDA. 2) Limited spatial resolution: we observed that AnyOKP performed better when object scale was large in image, and relatively worse when object scale was too small. 3) Requirement to in-plane rotation and object spacing: if an object is relatively rotated around the optical axis more than 90° or overlapped with another object with the same category, the method probably will give a unsatisfying result.

## V. CONCLUSION

In this paper, a novel AnyOKP approach is proposed for flexible object-centric visual perception. By introducing the training-free feature enhancement, BPP-based keypoint extraction, and edge-based keypoint grouping methods, we extend the traditional single-instance keypoint extraction to the multi-instance keypoint extraction, which is meaningful for many robot and automation applications. Exploiting the cutting-edge pretrained ViT as feature extractor, AnyOKP demonstrates remarkable robustness to domain shift and viewpoint change. In the future work, we will attempt to improve the real-time performance and spatial resolution.

## REFERENCES

[1] P. Rosenberger, A. Cosgun, R. Newbury, *et al*., "Object-independent human-to-robot handovers using real time robotic vision," *IEEE Robot. Autom. Lett.*, vol. 6, no. 1, pp. 17-23, 2021.

[2] F. Qin, D. Xu, D. Zhang, *et al*., "Robotic skill learning for precision assembly with microscopic vision and force feedback," *IEEE/ASME Trans. Mechatronics*, vol. 24, no. 3, pp. 1117-1128, 2019.

[3] Z. Zeng, *et al*., "Semantic linking maps for active visual object search," in *Proc. IEEE Int. Conf. Robot. Autom.*, 2020, pp. 1984-1990.

[4] S. Lin, F. Qin, H. Peng, *et al*., "Multi-frame feature aggregation for real-time instrument segmentation in endoscopic video," *IEEE Robot. Autom. Lett.*, vol. 6, no. 4, pp. 6773-6780, 2021.

[5] R. Xu, F.-J. Chu, C. Tang, *et al*., "An affordance keypoint detection network for robot Manipulation," *IEEE Robot. Autom. Lett.*, vol. 6, no. 2, pp. 2870-2877, 2021.

[6] J. Wang, S. Lin, C. Hu, *et al*., "Learning semantic keypoint representations for door opening manipulation," *IEEE Robot. Autom. Lett.*, vol. 5, no. 4, pp. 6980-6987, 2020.

[7] Z. Qin, K. Fang, Y. Zhu, *et al*., "KETO: Learning keypoint representations for tool manipulation," in *Proc. IEEE Int. Conf. Robot. Autom.*, 2020, pp. 7278-7285.

[8] Y. Lin, J. Tremblay, S. Tyree, *et al*., "Single-stage keypoint-based category-level object pose estimation from an RGB image," in *Proc. IEEE Int. Conf. Robot. Autom.*, 2022, pp. 1547-1553.

[9] R.-Q. Li, X.-L. Xie, X.-H. Zhou, *et al*., "Real-time multi-guidewire endpoint localization in fluoroscopy images," *IEEE Trans. Med. Imag.*, vol. 40, no. 8, pp. 2002-2014, 2021.

[10] J. Lu, F. Richter and M. C. Yip, "Pose estimation for robot manipulators via keypoint optimization and Sim-to-Real transfer," *IEEE Robot. Autom. Lett.*, vol. 7, no. 2, pp. 4622-4629, 2022.

[11] D. Hadjivelichkov, S. Zwane, *et al*., "One-shot transfer of affordance regions? affcorrs!," in *Proc. Conf. Robot Learn.*, 2023, pp. 550-560.

[12] K. Koreitem, F. Shkurti, T. Manderson, *et al*., "One-shot informed robotic visual search in the wild," in *Proc. IEEE/RSJ Int. Conf. Intell. Robots Syst.*, 2020, pp. 5800-5807.

[13] F. Qin, D. Xu, B. Hannaford, *et al*., "Object-agnostic vision measurement framework based on one-shot learning and behavior tree," *IEEE Trans. Cybern.*, vol. 53, no. 8, pp. 5202 - 5215, 2023.

[14] Y. Zhao, X. Guo and Y. Lu, "Semantic-aligned fusion transformer for one-shot object detection," in *Proc. IEEE/CVF Conf. Comput. Vis. Pattern Recognit.*, 2022, pp. 7601-7611.

[15] W. Jiang, E. Trulls, J. Hosang, *et al*., "Cotr: Correspondence transformer for matching across images," in *Proc. IEEE/CVF Int. Conf. Comput. Vis.*, 2021, pp. 6207-6217.

[16] P. E. Sarlin, D. DeTone, T. Malisiewicz, *et al*., "Superglue: Learning feature matching with graph neural networks," in *IEEE/CVF Proc. Conf. Comput. Vis. Pattern Recognit.*, 2020, pp. 4938-4947.

[17] C. Zhong, C. Yang, F. Sun, *et al*., "Sim2real object-centric keypoint detection and description," in *Proc. AAAI Conf. Artificial Intell.*, 2022, pp. 5440-5449.

[18] D. DeTone, *et al*., "SuperPoint: Self-supervised interest point detection and description," in *Proc. IEEE/CVF Conf. Comput. Vis. Pattern Recognit. Workshops*, 2018, pp. 337-349.

[19] F. Qin, J. Qin, *et al*., "Contour primitive of interest extraction network based on one-shot learning for object agnostic vision measurement," in *Proc. IEEE Int. Conf. Robot. Autom.*, 2021, pp. 4311–4317.

[20] F. Qin, D. Xu, D. Zhang, *et al*., "Automated hooking of biomedical microelectrode guided by intelligent microscopic vision," *IEEE/ASME Trans. Mechatronics*, online published, 2023.

[21] M. Dusmanu, I. Rocco, T. Pajdla, *et al*., "D2-Net: A trainable CNN for joint description and detection of local features," in *Proc. Conf. Comput. Vis. Pattern Recognit.*, 2019, pp. 8092-8101.

[22] J. Sun, Z. Shen, Y. Wang, *et al*., "LoFTR: Detector-free local feature matching with transformers," in *Proc. IEEE/CVF Conf. Comput. Vis. Pattern Recognit.*, 2021, pp. 8922-8931.

[23] S. Amir, Y. Gandelsman, S. Bagon, *et al*., "Deep ViT features as dense visual descriptors," arXiv preprint arXiv:2112.05814, 2(3), 4.

[24] K. Aberman, J. Liao, *et al*., "Neural best-buddies: sparse cross-domain correspondence," *ACM Trans. Graph.*, vol. 37, no. 4, Article 69, 2018.

[25] K. He, X. Zhang, S. Ren, *et al*., "Identity mappings in deep residual networks," in *Proc. Europ. Conf. Comput. Vis.*, pp. 630-645, 2016.

[26] A. Dosovitskiy, L. Beyer, A. Kolesnikov, *et al*., "An image is worth 16x16 words: Transformers for image recognition at scale," in *Proc. Int. Conf. Learn. Represent.*, 2021.

[27] M. M. Naseer, K. Ranasinghe, S. J. Khan, *et al*., "Intriguing properties of vision transformers," in *Proc. Advances Neural Inf. Process. Syst.*, 2021, pp. 23296-23308.

[28] M. Raghu, T. Unterthiner, S. Kornblith, *et al*., "Do vision transformers see like convolutional neural networks?," in *Proc. Advances Neural Inf. Process. Syst.*, 2021, pp. 12116-12128.

[29] K. Chen, S. Wang, B. Xia, *et al*., "TODE-Trans: Transparent object depth estimation with Transformer," in *Proc. IEEE Int. Conf. Robot. Autom.*, 2023, pp. 4880-4886.

[30] I. Radosavovic, T. Xiao, *et al*., "Real-world robot learning with masked visual pre-training," in *Proc. Conf. Robot Learn.*, 2023, pp. 416-426.

[31] A. Vaswani, N. Shazeer, N. Parmar, *et al*., "Attention is all you need," in *Proc. Advances Neural Inf. Process. Syst.*, 2017.

[32] A. Kirillov, E. Mintun, E., N. Ravi, *et al*., "Segment anything", *arXiv preprint arXiv:2304.02643*, 2023.

[33] M. Caron, H. Touvron, I. Misra, *et al*., "Emerging properties in self-supervised vision transformers," in *Proc. IEEE/CVF Int. Conf. Comput. Vis.*, 2021, pp. 9650-9660.

[34] M. Oquab, T. Darcet, *et al*. "Dinov2: Learning robust visual features without supervision," *arXiv preprint arXiv:2304.07193*, 2023.